\documentclass[11pt,a4paper]{article}
\usepackage[hyperref]{acl2019}
\usepackage{times}
\usepackage{latexsym}
\usepackage{bm}

\usepackage{url}

\aclfinalcopy 

\def\aclpaperid{183} 

\title{Sentence Level Curriculum Learning for Improved Neural Conversational Models}

\author{ Sean Paulsen \\
  Dartmouth College \\
  \texttt{sean.d.paulsen.GR@dartmouth.edu} \\}

\date{}

\begin{document}
\maketitle
\begin{abstract}
 Designing machine intelligence to converse with a human user necessarily requires an understanding of how humans participate in conversation, and thus conversation modeling is an important task in natural language processing. New breakthroughs in architecture and data gathering continue to push the performance of such conversational AI models. However, designs neglect the gradual buildup in sentence structure and complexity experienced by humans as we learn to communicate. During training, our model accepts one or more sentences as input and attempts to predict the next sentence in the conversation one word at a time, so our goal is to separate training into segments, with each segment's corpus comprised of longer sentence pairs than the previous one. This will mimic the desired ``buildup'' component of human learning. We begin with \textit{only} ``short'' length sentence pairs, then \textit{only} ``medium'' length pairs, and so on. A majority of our experiments were toward optimizing this technique, ensuring a proper representation of the technique's potential, since many of the details were new questions. Our segment-trained models were then able to achieve lower validation loss at the end of training than models trained with standard text preparation. This segmented training is straightforward to implement and our results provide a general direction for future research to implement and improve it.  
   \end{abstract}

\section{Introduction}
We take inspiration from the work done by \cite{Vogelsang:17} which adapts the physical process of humans learning to see to a CNN machine vision model. Newborns have lower visual acuity, so our brains begin learning to see using degraded (i.e blurry) images. As demonstrated in their work, this degradation is actually critical for learning to process configural face judgments, a fact reflected in the inability of older children who are treated for congenital cataracts to properly perform that facial analysis. \cite{Vogelsang:17} hypothesized that the same benefit could be found in the receptive fields and performance of a convolutional neural network being trained for machine vision. ``The results show that commencing training with blurred images creates receptive fields that integrate information across larger image areas and leads to improved performance and better generalization across a range of resolutions.'' \cite{Vogelsang:17} \cite{Bengio:2009} had established a more general treatment of this process that they called "Curriculum Learning," which has now become the standard terminology, and is the term used in this paper, and abbreviated as CL when the context is clear. We hypothesized that a curriculum of increasing sentence complexity would result in higher performance on next-utterance prediction.

Humans do not learn to participate in conversation by attempting to respond to sentences of random length. Our parents give us simple statements and questions to respond to, and our responses begin as statements with very few words and little complexity. However this is not reflected in the training regimens of modern conversational models. Similar length sentences are generally grouped together within a batch, but batch randomization disrupts any sense of gradual learning. Thus our analogue of ``degraded images,'' that is, our \textit{curriculum}, is to train in three segments, where the training data sentence pairs have the same ``length'' in each segment, length being either ``short,'' ``medium,'' or ``long'' (this is discussed in greater detail in \textbf{Section 3}). Our overall goal was to compare the performance of a model trained with this sentence level curriculum learning to a model trained on a standard all-inclusive, disordered dataset, but the finer details of such a process were a mystery as we could not find prior work on this technique at the sentence level. \textbf{Section 3} explores these mysteries and our approach to solving them. 

Afterwards, we compile several different test sets and compare the final validation loss of our curriculum learning models to those of two control group models. The first is trained on a corpus consisting in thirds of pairs from the short, medium, and long data, all mixed together randomly. This is to confirm that the ordering of increasing complexity contributes to overall performance.The second control model has no length-based preprocessing other than the standard general grouping by length within each batch. That is, each sentence pair in this control model's training data could have either sentence be of any length. This model represents the standard approach to training conversational agents. 

We needed a consistent and approachable conversational architecture as a baseline, and found a perfect candidate in \cite{Vinyals:15}. Our hyperparameters were optimized via random search \cite{Bergstra:12}, and can be found in the supplemental materials, along with a sketch of our Keras implementation of the model. We defer to \cite{Vinyals:15} for the remaining architectural details unless otherwise specified. 

\section{Related Work}
The last few years have seen the emergence of many end-to-end architectures for conversational agents and NLP in general. A review of ``significant deep learning related models $\ldots$ employed for numerous NLP tasks,''\cite{Young:17} showed that these conversational models share overall similarities with \cite{Vinyals:15}, yet none of them have addressed the gradual built-up nature of human speech that our work attempts to capture. Variations of the Transformer \cite{Vaswani:17} architecture are currently responsible for state of the art conversational agents. Indeed, ``pretrained Transformer variants are currently the best performing models on this task."\cite{Dinan:19} Yet even these models continue to train on both simple and complex data samples within the same regimen. \cite{Hancock:19}, for example, remarks that ``dialogue agents would ideally learn directly from dialogue $\ldots$ corresponds to the way humans learn to converse,'' similar to our desire to mimic a human learning process. Their models, like ours, are first trained on the familiar ``next-utterance prediction'' task, using the PersonaChat dataset \cite{Zhang:18}. However, their corpus of next-utterance training pairs include samples such as ``How are you?'' ${\rightarrow}$ ``Great, thanks!'', but also ``I do not have children at the moment.'' $\rightarrow$ ``That just means you get to keep all the popcorn for yourself.''\cite{Zhang:18} No regard is then given to the significant difference in semantic complexity between the pairs. Our initial work on this topic has reduced that notion of ``semantic complexity'' to merely ``length,'' but we look forward to refining the definition of our ``degraded samples'' in future work. \cite{Bengio:2009} examined a curriculum of increasing vocabulary size on a task of next-\textit{word} prediction, which showed statistically significant improvement. Their curriculum and task were not the only important differences from our work, however. Their training corpus consisted of windows of text from Wikipedia, so there is no notion of the conversational aspects we are focused on, and no heed was paid to the complexity of the text itself. 

\section{A Sentence Level Curriculum}
We need to optimize the our curriculum before we can draw conclusions about models that are trained according to it. More specifically, we need to answer these two questions: $1)$ How are ``short,'' ``medium,'' and ``long'' sentences defined? $2)$ How many pairs of each length should we train on, and for how long?

\noindent  Due to time constraint, we could not perform a proper grid search to optimize the lengths of the three segments, and thus we somewhat arbitrarily defined a ``short sentence'' as being between 1 and 4 (inclusive) words, not counting the start-of-sentence and end-of-sentence tokens added during preprocessing, a ``medium sentence'' as 5 to 10 words, and a ``long sentence'' as 10 to 16 words. Define a \textbf{length pair} to then be comprised of two successive sentences of dialogue appearing somewhere in the OpenSubtitles corpus. Note that our task does not predict the utterance of the next \textit{speaker}, merely the next utterance in general, and so it may be that both sentences were spoken by the same person. This is a flaw we choose to tolerate for the time being. The first bullet above is resolved. 

After processing the full OpenSubtitles english dataset\cite{Vinyals:15}, we have 68 million short length pairs, 59 million medium length pairs, and 7 million long length pairs. However, due to hardware and time limitations, the maximum training a given model can receive on a given segment is 3 million samples (with a batch size of 128) for 6 epochs, which brings us to the second bullet. Do we really want the full treatment on each segment? Our intuition was that more data and more time (until overfitting begins) should be superior. However, note that when the model trains on short length pairs, it is going to learn very quickly that the end-of-sentence token, ``$<$eos$>$'', comes after only a few words, but more importantly, that it \textit{never} takes more than four words to reach the end of the sentence. Then once we begin training on the medium segment, the model will have a difficult time predicting anything other than ``$<$eos$>$'' after four words, and that prediction will be incorrect in every training sample. Perhaps the additional learning from additional training time on the short segment is not worth the trouble of correcting a more ironclad notion of where and when to predict ``$<$eos$>$''. We have called this potential problem "overspecializing," where the model becomes \textit{too} familiar with the ``degraded'' samples and is therefore overly resistant to learning ``higher resolution'' samples. Overspecialization could theoretically be caused by training on too many samples, by spending too many epochs on those samples, or both. Thus the problem of optimizing our curriculum has now been reduced to minimizing overspecialization. On one end, if overspecialization is simply not real, then we expect to see the models trained on the most data for the most time to be the most effective learners in the medium segment, that is, to end medium segment training with the lowest loss values, and we expect the analogous result if indeed overspecialization is a crippling problem. 

We gathered training sets of sizes 10K, 50K, 200K, 500K, 1M, and 3M short length pairs, chosen uniformly at random across the 68 million available pairs. To account for the impact of random initial values of parameters, we trained ten models on each of the six sample sizes, and all were trained for 6 epochs. But recall that we're interested in the extent of overspecialization due to training time, so we saved all the parameters of each model after 2, 4, and 6 epochs. This resulted in 180 sets of parameters to consider. Unwieldy, but accounting for initial values contributed a factor of 10. So for each of the 18 dataset/epoch combinations, we search the ten results for the lowest validation loss and declare these sets of weights the ``winners.'' The 18 winners' short-trained weights are then inherited by a new model to be trained on the medium segment.

But we have the same problem at this stage, how much of the available medium training data should we use and how many epochs should we train for? We gathered the same six sizes of training data, medium pairs chosen uniformly at random from the 59 million available pairs, and trained all combinations of medium training data and short-winners for six epochs, for a total of 108 models and 324 sets of short-medium-trained parameters. These results granted us preliminary evidence of overspecialization in our models. To see this, consider a hypothetical experiment where our overall task is only to predict next-utterance for medium length pairs after training on short length pairs. Then we would have our many inherited parameters after short training as we do now, and we would train on as much medium data as we can for as long as possible. So we can look at our results so far for that specific case, that is, models trained on 3 million medium length pairs for 6 epochs. Then, given a short-training data size, which number of epochs on that short length data resulted in the lowest loss value? Table 1 shows this information. Indeed the best loss overall for 3 million/6 epochs medium training had short segment hyperparameters in the middle, 200K short pairs for only 2 epochs. More specifically, for the task of predicting medium length next-utterances, \textbf{a pre-training regimen of 200K short length pairs for 2 epochs was superior to other ``intuitively superior'' configurations such as 200K for 6 epochs, 3M for 2 epochs, and 3M for 6 epochs}. This is good evidence to support our concerns about overspecialization. It is important to note that our models are programmed for validation based early stopping, which is to say that the validation accuracy improved with each epoch during short-training, so this is fundamentally different from overfitting. 

\begin{table}[t!]
\centering
\begin{tabular}{llllll}
\textbf{Short TD} & \textbf{Short Epochs} & \textbf{Val. Loss}\\
\hline 10000 & 6& 3.786\\
\hline 50000 & 2 & 3.785\\
\hline 200000 & 2 & 3.782\\
\hline 500000 & 4 & 3.799\\
\hline 1000000 & 2 & 3.788\\
\hline 3000000 & 2 & 3.791\\
\end{tabular}
\caption{Given a short-segment training set size, we identified the best training performance on the 3 million medium length pairs after 6 epochs and checked how many epochs it spent training on that short length data.}
\end{table}

Ideally we would have liked to continue in this full grid-search manner, training models initialized with all 324 short-medium-trained parameters on all six different sizes of long training sets, but this was simply not feasible in time or hardware. Thus one-sixth of the 324 sets of parameters were chosen uniformly at random to be inherited by the models in the long segment, which, in the interest of time, had training sets only of size 1 million and 3 million, and all models were trained for the full six epochs. Two of these models encountered errors in our high performance cluster during training and were lost. Table 2 shows the results. The three additional ``types'' below the curriculum learning results are explained in \textbf{Section 4}, and were included in this table for cleanliness. 

As we have come to expect, the best performance was not trained for 6 epochs on either medium or short length data, and the worst performances did not have enough exposure to medium length pairs with the smaller medium training set sizes, even with 6 epochs to learn from them. We see substantial improvement on the 3 million large-length set when reducing the short-length set from 3 million down to 50 thousand, while the full 3 million medium set is common between them. This is likely do to the semantic similarity of our ``medium'' and ``large'' data. The notion of separate clauses or other abstract pieces of a sentence is gained in the medium segment, since the short sentences are some single, simple clause. So training on more medium data exposes the model to more variations of how those abstract pieces can fit together, which helps to learn the long data, but only for 4 epochs so as not to become too attached to the more immediate placement of the end-of-sentence token, as well as whatever unknown contributors there are to overspecialization. This was of course different from what we saw when moving from short to medium, because the difference in semantics was more significant, i.e single clause to multi clause is more significant than multi clause to slightly longer multi clause. 

\begin{table*}[t!]
\centering
\begin{tabular}{lllllllllll}
\textbf{Type} & \textbf{Val. Loss} & \textbf{Long TD} & \textbf{Med TD} & \textbf{Med Epochs} & \textbf{Short TD} & \textbf{Short Epochs}\\

\hline DT (worst) & & & & & & \\
\hline & 3.941 & 1000000 & 10000 & 6 & 200000 & 2\\
 \hline& 3.904 & 1000000 & 10000 & 2 & 10000 & 6\\
\hline & 3.878 & 1000000 & 10000 & 4 & 500000 & 6\\ 
\hline \\
\hline & $3.618$ & 3000000 & 10000 & 6 & 10000 & 4\\
\hline & 3.578 & 3000000 & 50000 & 6 & 500000 & 4\\
\hline & 3.574 & 3000000 & 10000 & 6 & 500000 & 4\\
\hline DT (best) & & & & & & \\
 \hline& 3.783 & 1000000 & 1000000 & 4 & 50000 & 2\\
\hline & 3.788 & 1000000 & 3000000 & 6 & 10000 & 2\\
\hline & 3.791 & 1000000 & 1000000 & 4 & 1000000 & 4\\
\hline \\
\hline & 3.473 & 3000000 & 3000000 & 4 & 50000 & 4\\
\hline & 3.483 & 3000000 & 3000000 & 4 & 3000000 & 4\\
 \hline& 3.485 & 3000000 & 3000000 & 6 & 200000 & 6\\
 
 \hline
Fresh & & & & & & \\
\hline & $4.028$ & 1000000 & & & &\\
\hline & $3.705$ & 3000000 & & & &\\

\hline
Mix & & & & & & \\
\hline & 4.199 & 1000000 & & & & \\
\hline & 4.188 & 3000000 & & & & \\

\hline
Cross & & & & & & \\
\hline & 4.625 & 1000000 & & & & \\
\hline & 4.577 & 3000000 & & & & \\

 \end{tabular}
  \caption{The lowest validation loss for fresh-long, cross, and mix models on the 1 million and 3 million sized long length validation sets for comparison against the best and worst performers on long length training sets after curriculum learning.}
 \end{table*}
 
\section{Results}

An immediate question to ask at this point is whether our curriculum learning has accomplished anything at all. Perhaps it was optimal to train on short length pairs for short periods of time because that made it easier to \textit{forget} what was learned, so the medium segment training approximated training with fresh, randomly initialized models. A quick experiment yields evidence to the contrary. We trained five fresh models on a training set of 1 million long length pairs, and five more on 3 million, all for the full six epochs, then looked at the best performance for the two training sets. These are the ``fresh'' type models included in Table 2 below. 

As the table shows, even the worst performing curriculum trained models outperformed the fresh models. Gaining advantage from pre-training is not surprising in and of itself, but in our case what was effectively pre-training consisted of two training sets that are fundamentally different from each other, as well as from the current training set. The model must have learned from the short and medium segments some things about dialogue that are not necessarily specific to the length of the samples in order to have obtained this edge in training on the long segment. It is worth noting, however, that the fresh models on 3 million long pairs were able to out perform the curriculum trained models that only had 1 million long pairs to learn from. Now how do these compare against models trained with standard training sets?

Define \textbf{mix pairs} to be a training set comprised one-third each of short, medium, and long length pairs, chosen uniformly at random from their respective collections, and ordered randomly within this mix pairs collection. So each pair has two sentences of the same length, but the set contains pairs of all three lengths. This is the natural training set to acquire models for comparison against our curriculum training, that is, to determine if the segmentation had the desired effect. Define \textbf{cross pairs} to be a training set drawn from OpenSubtitles with no restrictions other than the maximum length of either sentence within a pair remains at 16. Models trained on cross pairs represent the standard method of training for us to compare against. We train five models on 1 million cross pairs for 6 epochs, and five more on 1 million mix pairs for 6 epochs, and choose the best of each to account for random initialization. These two winners are the representatives of the cross and mix ``types'' in Table 2. The ``Long TD'' column for these models refers to the validation set that was used for evaluation.

Even the worst performing curriculum learning models performed better than the three comparison types on a long-length validation set by a considerable amount. Notice, though, that the fresh model representative outperformed the mix and cross models as well, due to training specifically on the task being evaluated, i.e long pair next-utterance prediction. Having just completed training on its long segment, the curriculum trained model has an advantage over the mix and cross models in this evaluation for that same reason.Therefore it is difficult to quantify the improvement due only to the curriculum from these results alone. Further, a quick check shows our best curriculum trained models performing much worse than the standard models on a test set of mix pairs, by about the same margin that they are superior in Table 2. So if we want to make this sort of direct comparison, we need to be more deliberate in designing our curriculum toward that goal. For example, proportions of successive segments contain samples from previous segments so that the final segment is approximately a mix pairs training set. \textit{That} would serve for a direct comparison against the mix model we trained here, and is closer to answering the question we set out with. This is expanded upon in the following section.

For now, though, having taken merely our first step into the territory of sentence level curriculum learning, these are positive results. We hypothesized the influence of overspecialization and observed strong evidence of its existence both when moving from short to medium and from medium to large. Further, we observed that short and medium segment training drastically improved performance on long pairs compared to a fresh model, despite all the short and medium pairs being outside the domain of the long segment's task.

\section{Future Work}
There are many important questions remaining. How would our results change by varying the cutoffs in between segments, or perhaps by adding more segments? Our analogue to different resolutions of images was different lengths of sentence pairs, while a more fitting analogue would have some notion of overall semantic complexity. How exactly to define and measure such a quantity remains an open problem. Further, similar to length, what would be the cutoffs in this complexity value from segment to segment? The answer might only be found experimentally, and similar to our experience, optimizing the hyperparameters of such a regimen would be quite taxing. 

As mentioned near the end of the previous segment, one glaring flaw with our approach is that humans do not stop using short sentences in conversation merely because they've learned to use medium sentences, and so on for long sentences. How would our results change if each segment included not only length pairs of the current segment length, but length pairs, mix pairs, and cross pairs of current and all previous segments? As always, tuning the proportions of each type of pair that would comprise such a dataset would be a significant experimental endeavor. 

Most imminent, though, is to run these same experiments on state of the art Transformer variants. Doing so is straightforward with the model in hand, one need only divide their training data into segments and tune the regimen, ideally with a more deliberate plan toward producing evidence of overspecialization, now that we have a better idea of how to demonstrate it. Per \cite{Dinan:19}, PersonaChat seems to result in better conversational performance than OpenSubtitles with these models. But, as mentioned near the end of \textbf{Section 2}, it remains to be seen if PersonaChat is large enough to survive being segmented. 

\bibliographystyle{acl_natbib}
\bibliography{degradation}

\end{document}